\documentclass[final,nonatbib]{article}

\usepackage[final]{neurips_2019}

\usepackage[utf8]{inputenc} 
\usepackage[T1]{fontenc}    
\usepackage{hyperref}       
\usepackage{url}            
\usepackage{booktabs}       
\usepackage{amsfonts}       
\usepackage{nicefrac}       
\usepackage{microtype}      
\usepackage{eqnarray,amsmath}
\usepackage{subfig}
\usepackage{graphicx}
\usepackage{authblk}
\title{Intelligent Coordination among Multiple Traffic
Intersections Using Multi-Agent Reinforcement Learning}

\usepackage{authblk}

\author[1]{\textbf{Ujwal Padam Tewari}}
\author[2]{\textbf{Vishal Bidawatka}}
\author[1]{\textbf{Varsha Raveendran}}
\author[1]{\textbf{Vinay Sudhakaran}}
\author[3]{\textbf{Kodate Shreedhar Shreeshail}}
\author[3]{\textbf{Jayanth Prakash Kulkarni}}
\affil[1]{\textbf{Advanced Data Management Research Group, \centerline{ Siemens Technology and Services Pvt. Ltd., Bengaluru, India}}      \textit{\{ujwal.tewari, varsha.raveendran, vinay.sudhakaran\}@siemens.com}}
\affil[2]{\textbf{International Institute of Information Technology Hyderabad}
\textit{\centerline{vishal.bidawatka@students.iiit.ac.in}}}
\affil[3]{\textbf{Indian Institute of Science Bengaluru}
\textit{\centerline{krishna.shreedhar@gmail.com, kulka112@uwindsor.ca}}}

\begin{document}
\maketitle
\begin{abstract}

  \par  We use Asynchronous Advantage Actor Critic (A3C) for implementing an AI agent in the controllers that optimize flow of traffic across a single intersection and then extend it to multiple intersections by considering a multi-agent setting. We explore three different methodologies to address the multi-agent problem - \textbf{(1)} use of asynchronous property of A3C to control multiple intersections using a single agent   \textbf{(2)} utilise self/competitive play among independent agents across multiple intersections and \textbf{(3)} ingest a global reward function among agents to introduce cooperative behavior between intersections. We observe that \textbf{(1)} \& \textbf{(2)} leads to a reduction in traffic congestion. Additionally the use of \textbf{(3)}  with \textbf{(1)} \& \textbf{(2)} led to a further reduction in congestion.
\end{abstract}

\section{Introduction}
Increasing vehicle population worldwide has led to traffic congestion, delays and unpredictable travel times. Existing adaptive traffic control systems (ATCS) are widely used to select traffic signal plans based on a real-time model of the traffic. Currently traffic engineers design signal plans and hand-tune the necessary parameters for different traffic situations to ensure optimal flow of traffic. It is, however, difficult to manually design signal plans for a large number of traffic scenarios, especially in the event of traffic anomalies such as accidents, road construction and traffic violations such as illegal parking/stopping. Traditional control algorithms work under predefined rules and suffer from high inductive bias. Typically, such bias is supplied by domain experts and require manual intervention. 

Our goal is to control traffic signals in an adaptive manner by providing a variable green signal time,  based on instantaneous traffic flow without the need for hand engineered signal plans. We use the asynchronous property of  A3C\cite{mnih2016asynchronous} algorithm to build single and multi-agent reinforcement learning systems. These algorithms learn the local and global optimal behavior of multiple intersections such that overall traffic flow is maximized. A global reward function is defined to incorporate cooperation among multiple agents to optimize global policies.  

The remainder of the paper is organized into the following sections: Section \ref{refs} presents related work. Section \ref{setting} introduces the simulation environment along with state space, action space and reward functions. In Section \ref{algorithm} we discuss the different algorithms evaluated for single and multi-agent network architectures. Section \ref{result} describes in detail the results and conclusions.

\section{Related Work} \label{refs}

The work done in \cite{liang2018deep} provides insights to a basic state encoding of the traffic signal. Studies such as \cite{genders2016using} and \cite{gao2017adaptive} use Deep Q-Networks with experience replay\cite{adam2011experience} and target network\cite{lillicrap2015continuous} to address sampling\cite{zhai2016deep} and non-stationary target\cite{foerster2016learning} issues respectively. When dealing with a large and continuous state space, methods like Monte Carlo\cite{guo2014deep}, DQN\cite{mnih2013playing}, DDQN\cite{tsividis2017human} are not efficient and face convergence and sampling problems \cite{pan2018organizing} \cite{arulkumaran2017deep}. Thus, in this work, we use Asynchronous Advantage Actor Critic-A3C \cite{mnih2016asynchronous}, a variant of actor-critic algorithms that uses the advantage function \cite{peters2008natural} to address convergence issues, in addition to  two separate actor and critic networks to handle sampling and non-stationary target problems. 

 \cite{tan1993multi} and \cite{littman1994markov}  explain the dynamics of multiple agents working together and independently in a cooperative environment to maximize their rewards. The team work by the players in \cite{OpenAI_dota} allow for collaboration among agents only by introducing a global reward function to induce team spirit among the operating players without an external linkage. \cite{zhang2018fully} proposes the use of actor-critic method in a multi-agent setup. 
 
 We also use the asynchronous property of A3C to deploy a single model onto multiple intersections, thereby reducing training time.

\section{Setting} \label{setting}
In this section we describe the simulation environment, state space, action space and reward functions used to train the RL (reinforcement learning) agent. 

\subsection{Environment}

\begin{figure}
     \centering
     \subfloat{\includegraphics[width=1\textwidth]{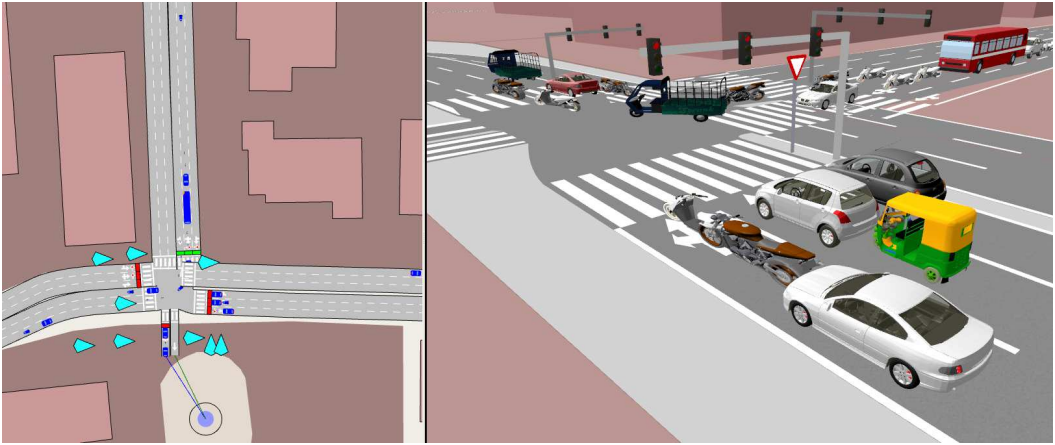}}
     \caption{Aimsun 2D and 3D simulation of traffic from the city of Bengaluru, India. Different composition of vehicles, including 2-wheelers and 3-wheelers are considered to create realistic simulations for accurate results.}
     \label{fig:aimsun rendering}
\end{figure}

We use Aimsun Next\cite{barcelo2005microscopic}\cite{casas2010traffic} to simulate realistic traffic conditions and train RL agents. The real traffic measurements (number of vehicles and traffic density) calculated by the approach described in \cite{yeshwanth2017estimation} is used to enhance the simulations. The collected data follows \textit{Weibull distribution}\cite{lai2003modified}. Fig.\ref{fig:aimsun rendering} shows a sample simulation in Aimsun Next.

\begin{figure}
     \centering
     
     \subfloat{\includegraphics[width=.8\textwidth]{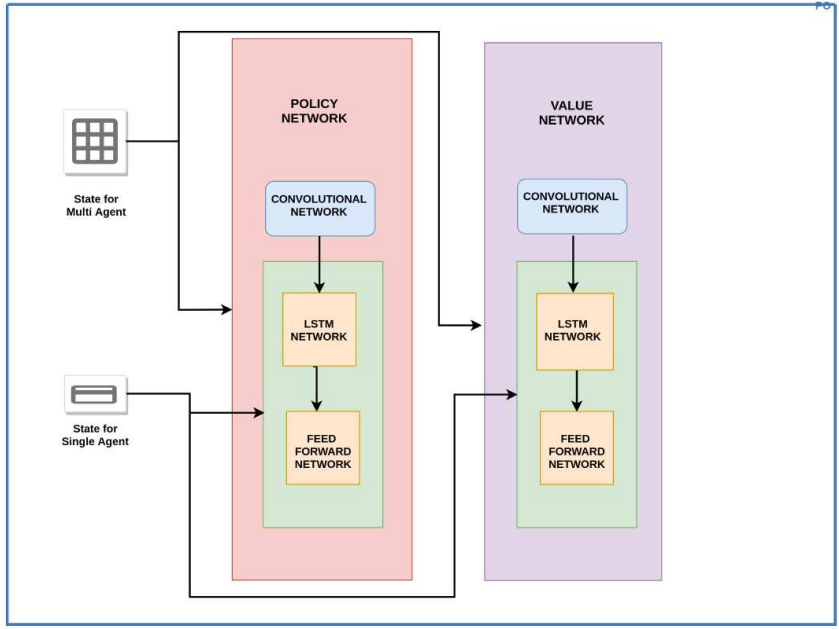}}
     \caption{Policy and Value Network used in A3C for both single and multi-agent setup. For the single agent the input is a single queue and for multi-agent setup it is a matrix, both containing density and phase information of single and multiple intersections respectively. Single agent utilises a Linear Layer + LSTM network while a multi-agent setup uses a Conv-LSTM architecture for its input. }
     \label{fig:a3c network used for single and multi}
\end{figure}

\subsection{State Space}
The state space of an agent controlling its intersection is \textbf{[encoded density, encoded phase]} where encoded density is the density of different approaches towards the intersection and encoded phase contains the intersection phase information. Density is defined as \textit{average number of vehicles per kilometer across all approaches of an intersection} and intersection phase information \textit{is the information encoding that represents sections of an intersection that are active (green) at an instance of time}. State encoding along with one-hot encoded phase information for a single agent is explained in Table \ref{Phase encoding}. 

The densities are normalized with the highest density among the approaches using eq.\ref{eq:density} and eq.\ref{eq:normalised}. This is done to help the model learn which section (or approach) has maximum amount of traffic and accordingly allocate necessary green time to it.


\begin{figure}
     \centering
     
     \subfloat[][Delay Graph]{\includegraphics[width=.5\textwidth]{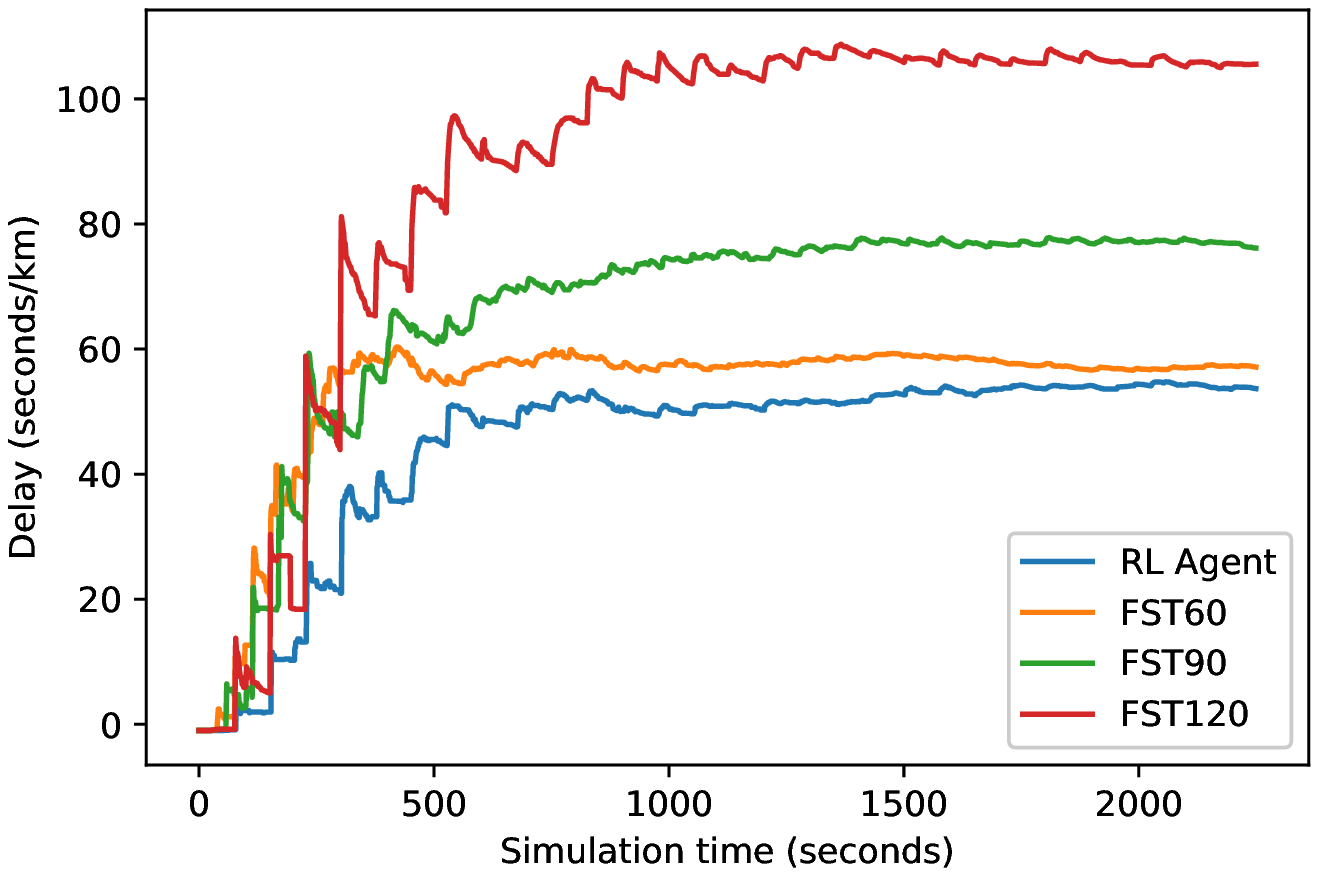}}
     \subfloat[][Density Graph]{\includegraphics[width=00.5\textwidth]{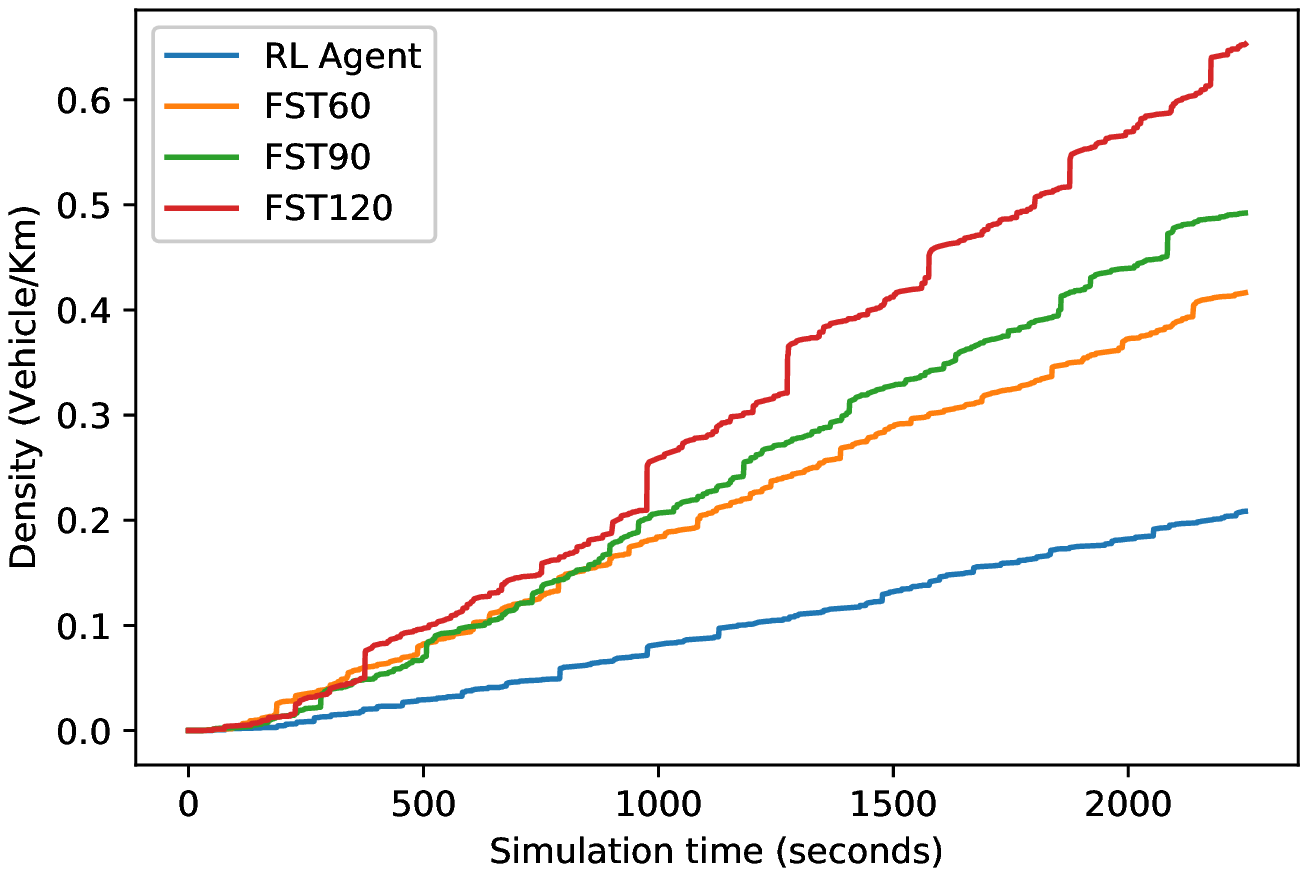}}
     \caption{ Comparison graphs for neighbouring intersection state space representation in a multi-agent scenario against Fixed Signal Timing (FST) of 60, 90 and 120 seconds. Delay and density on y-axis are the average delay (seconds per kilometer) and average density (vehicles per kilometer) across the four intersections respectively. X-axis is the simulation time in seconds.}
     \label{fig:state graphs}
     
\end{figure}

\begin{table}
  \caption{Single Agent State Space Encoding}
  \label{Phase encoding}
  \centering
  \begin{tabular}{llllll}
    \toprule
    \cmidrule(r){1-2}
    Phase     & Section Upper & Section Right & Section Lower & Section Left  
    & Encoded Phase \\
    \midrule
    1 & Green &Red &Red &Red  &[1,0,0,0]     \\
    2  &Red &Green &Red &Red  & [0,1,0,0]      \\
    3  &Red &Red &Green &Red  & [0,0,1,0]  \\
    4  &Red &Red &Red &Green  & [0,0,0,1]  \\
    \bottomrule
    \\
      & & & & &Encoded Density  \\
    \midrule
     Density &$D_1$ &$D_2$ &$D_3$ &$D_4$ &[$D_1$,$D_2$,$D_3$,$D_4$]\\
    \bottomrule
  \end{tabular}
\end{table}

\begin{equation}
    D_{max}  = max(D_1 ,D_2 ,D_3 ,D_4)
    \label{eq:density}
\end{equation}

\begin{equation}
    Encoded Density= \frac{[D_1,D_2,D_3,D_4]}{D_{max}}
    \label{eq:normalised}
\end{equation}

\begin{figure}
     \centering
     \subfloat{\includegraphics[width=.5\textwidth]{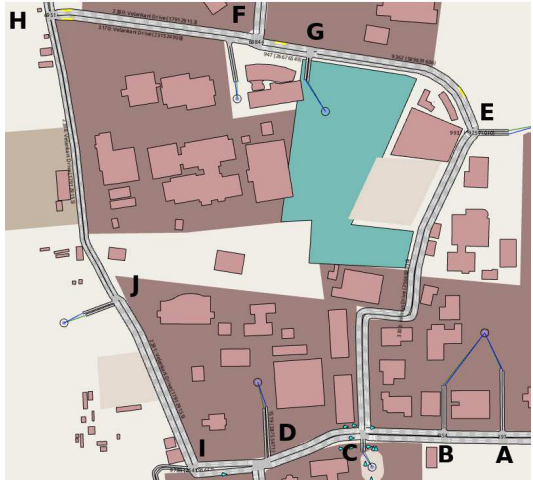}}
    
     \caption{Intersection representation in Aimsun of a small part of the city of Bengaluru in India for Neighbouring State Representation where each intersection is represented with an alphabet. }
     \label{fig:Neighbouring intersections in Aimsun}
\end{figure}

Figure \ref{fig:Neighbouring intersections in Aimsun} shows the road network considered to train RL agents in our experiments. In the figure, intersections D, B and E are neighbours of C. Their densities and phase information are considered as the state space for intersection C.  We represent the state space comprising of density and phase encoding of the neighboring intersections as a matrix shown below for training multi-agent networks.

$$ C= \left(\begin{matrix}C_{d1}&C_{d2}&C_{d3}&C_{d4}&C_{p1}&C_{p2}&C_{p3}&C_{p4}\\B_{d1}&B_{d2}&B_{d3}&B_{d4}&B_{p1}&B_{p2}&B_{p3}&B_{p4}\\E_{d1}&E_{d2}&E_{d3}&E_{d4}&E_{p1}&E_{p2}&E_{p3}&E_{p4}\\D_{d1}&D_{d2}&D_{d3}&D_{d4}&D_{p1}&D_{p2}&D_{p3}&D_{p4}\end{matrix}\right) $$

where $C_{d1} - C_{d4}$ represents the densities and $C_{p1} - C_{p4}$ represents the encoded phase information of the respective intersection. Similarly for intersection D, density and phase of intersections A and E are encoded in its state space. 

Fig.\ref{fig:state graphs} depicts the impact of using density and phase information of neighbouring intersections as the state space on average delay time\footnote{Delay time for a section in the intersection is calculated as average delay time per vehicle per kilometre. In Aimsun, it is the difference between the expected travel time (the time it would take to traverse the system under ideal conditions) and the travel time. For a section, it is calculated as the average of all vehicles and then converted into time per kilometre and does not include the time spent in a virtual queue.} and average density of intersections C, D, E and F as referred to in Fig. \ref{fig:Neighbouring intersections in Aimsun}.

\subsection{Action Space}

A discrete action space is considered with a fixed range from 20 to 60 seconds of green signal time in increments of +5. The model selects four different actions in the form of four different green signal times for each approach of the intersection. The action space is shown below-

Green signal time - [20, 25, 30, 35, 40, 45, 50, 55, 60]

We experimented with various action spaces such as green signal times from 0 to 60 seconds with increments of 5 and 10 seconds respectively and 20 to 120 seconds with increments of 20, yet none of them were effective in reducing the traffic congestion. The sum of the four green times obtained is always less than or equal to a certain threshold of the intersection which can be set manually. In the experiments performed, 240 seconds is used as the threshold.

\subsection{Reward Function}
The task of designing reward functions is important in solving a problem using reinforcement learning and therefore comprehensive testing for multiple reward functions is necessary. For our agents to not be solely governed by the reward functions\cite{bush2005modeling}\cite{glascher2010states}, we chose density (also a parameter in state-space) as our reward function parameter. The reward is estimated by considering densities of two different time stamps $D_{T}$ and $D_{T+1}$. We experimented with the following reward functions-
\begin{itemize}
\setlength\itemsep{.12em}
    \item \textbf{Product of Densities-} The product of densities over a particular time $T_{t}$ is calculated given in eq. \ref{eq:product}.
    
    \begin{align}
        D_{t} = \prod_{i=1}^{4} D^{i}_{t}
        \label{eq:product}
    \end{align}
    
    where $D^{i}$ is the density of the i$^{th}$ section of the intersection and $i$ ranges from 1 to 4 

    \item \textbf{Sum of Densities-} The sum of densities of each of the four sections of the intersection is calculated in eq.\ref{eq:sum}
    
    \begin{align}
        D_{t} = \sum_{i=1}^{4} D^{i}_{t}
        \label{eq:sum}
    \end{align}
    \item \textbf{Sum of Squares of Densities-} The sum of squares of density of each section is calculated in eq. \ref{eq:square}
    \begin{align}
        D_{t} = \sum_{i=1}^{4} {D^{i}_{t}}^{2}
        \label{eq:square}
    \end{align}
    
\end{itemize}

The final reward is calculated with the difference across two time-stamps as mentioned in equation \ref{eq:reward}. If the obtained reward is positive, a +1 is supplied, else if the reward is negative, a value of -1 is supplied. The clipping of rewards eliminate the possibility of sparse rewards, making the agent end goal oriented which in turn enhances the learning\cite{henderson2018deep}\cite{mnih2015human}.

 \begin{equation}
     Reward = D_{t}-D_{t+1}
        \label{eq:reward}
 \end{equation}

Among the reward functions mentioned above, the product of densities resulted in reducing the overall delay at the intersection. This is primarily because in a dense/chaotic environment, such as on Indian roads, the traffic is stochastic in nature. During peak hours the traffic densities are nearly the same and the product of densities will amplify the effect. The product as a reward function accommodates even the minute fluctuations in densities across four approaches in contrast to the sum and sum of squares of densities.


\section{Adaptive Signal Control using Single and Multi-agent architectures} \label{algorithm}

In the following section we elaborate on the techniques evaluated and showcase their results using average delay time and average density (in case of single agent) of intersections by taking FST (Fixed Signal Timing) as our baseline. In this method a fixed measure of green signal time is assigned to every approach of the intersection during each cycle in a simple round-robin fashion. The fixed signal timing of 60, 90 and 120 seconds is chosen for baseline evaluation of our RL based model since they are most frequently used as the traffic signal timings in Indian cities.
\ref{single agent} discusses single agent operating on an intersection. The remaining sections describe multi-agent behaviour to optimize traffic flow across multiple intersections.

\subsection{Single Agent - Single intersection}\label{single agent}

We evaluated a single agent controlling a single intersection by using real traffic data. The aim is to reduce the overall delay time at this intersection. The hyperparameters that yielded the best results in terms of lowest delay per vehicle at the intersection and the overall average density of intersection are given below -


\begin{itemize}
    \item Network - LSTM
    \item State Space - [Density | Phase] for each intersection
    \item Action Space - Green time: 20 seconds to 60 seconds with increments of 5 seconds
    \item Reward Function - Product of densities
\end{itemize}

\subsection{Independent RL (InRL)}
We utilize the notion of Independent RL wherein agents are trained collectively in the same environment and while doing so the agents formulate their own strategies of cooperation, without actually communicating with each other. This produces self competitive behaviour among the agents to maximize their individual cumulative reward while working together with other agents without any direct communication in place. The setup was imitated individually on four different agents operating across four different intersections which include A, B, C and D as shown in Fig.\ref{fig:Neighbouring intersections in Aimsun}.

\begin{figure}
     \centering
     
     \subfloat{\includegraphics[width=.8\textwidth]{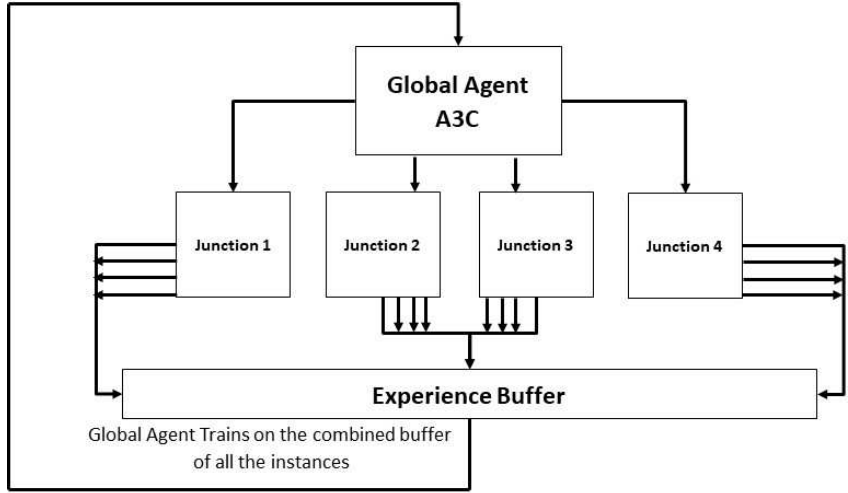}}

     \caption{Schematic Representation of A3C being used for multiple intersections. Each intersection is an asynchronous worker running in its individual environment. Each worker in the figure outputs: Current State, Reward, Action and Next State where State is Density and Action is the Green Signal Time.}
     \label{fig:a3c multi intersection}
\end{figure}

\subsection{Asynchronous Instances as intersections}
We now employ the asynchronous attribute of A3C where multiple instances of the environment are used for training, by deploying four different traffic intersections as four distinctive environments as illustrated in Fig.\ref{fig:a3c multi intersection}. Instead of having different agents for four different intersections, only one agent trained on the experience buffer of all four intersections decides the signal timings at their respective intersections.

\subsection{Coordination via Global Reward Function}
As we already know the performance of RL agents is governed by the reward function, we propose a model of global reward function for multi-agent setting. This global reward function fuses the agent's individual reward and also takes into account a global reward which is the average of rewards of all agents and is given in eq.\ref{eq:gloabl_reward}

\begin{align}
        Reward_{Global} = \frac{\sum_{i=1}^{4}{Reward_i}}{4} 
        \label{eq:gloabl_reward}
\end{align}
    
\begin{align}
    Reward_{Final} = 0.5* (Reward_{Global} + Reward_{Individual intersection}) 
    \label{eq:ireward}
\end{align}

The individual agent then takes the average value of the two reward functions - its own  and global reward. The final reward is clipped to +1 if positive and  -1 if negative. Currently we consider equal weight for both global as well as individual reward and intend to experiment with different ratios in the future.

\begin{figure}
     \centering
     
     \subfloat[][Delay Time comparison]{\includegraphics[width=.5\textwidth]{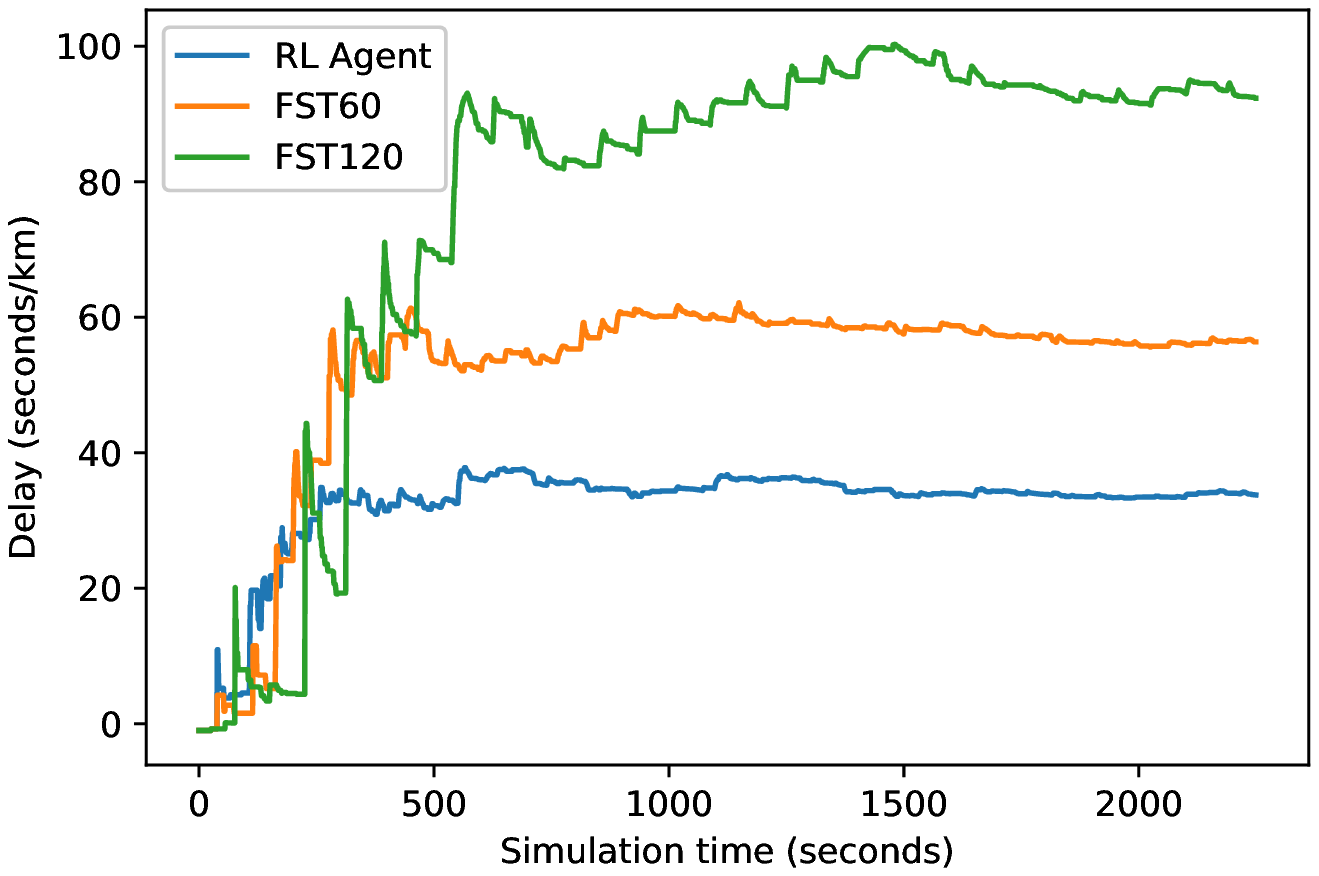}}
     \subfloat[][Density comparison]{\includegraphics[width=.5\textwidth]{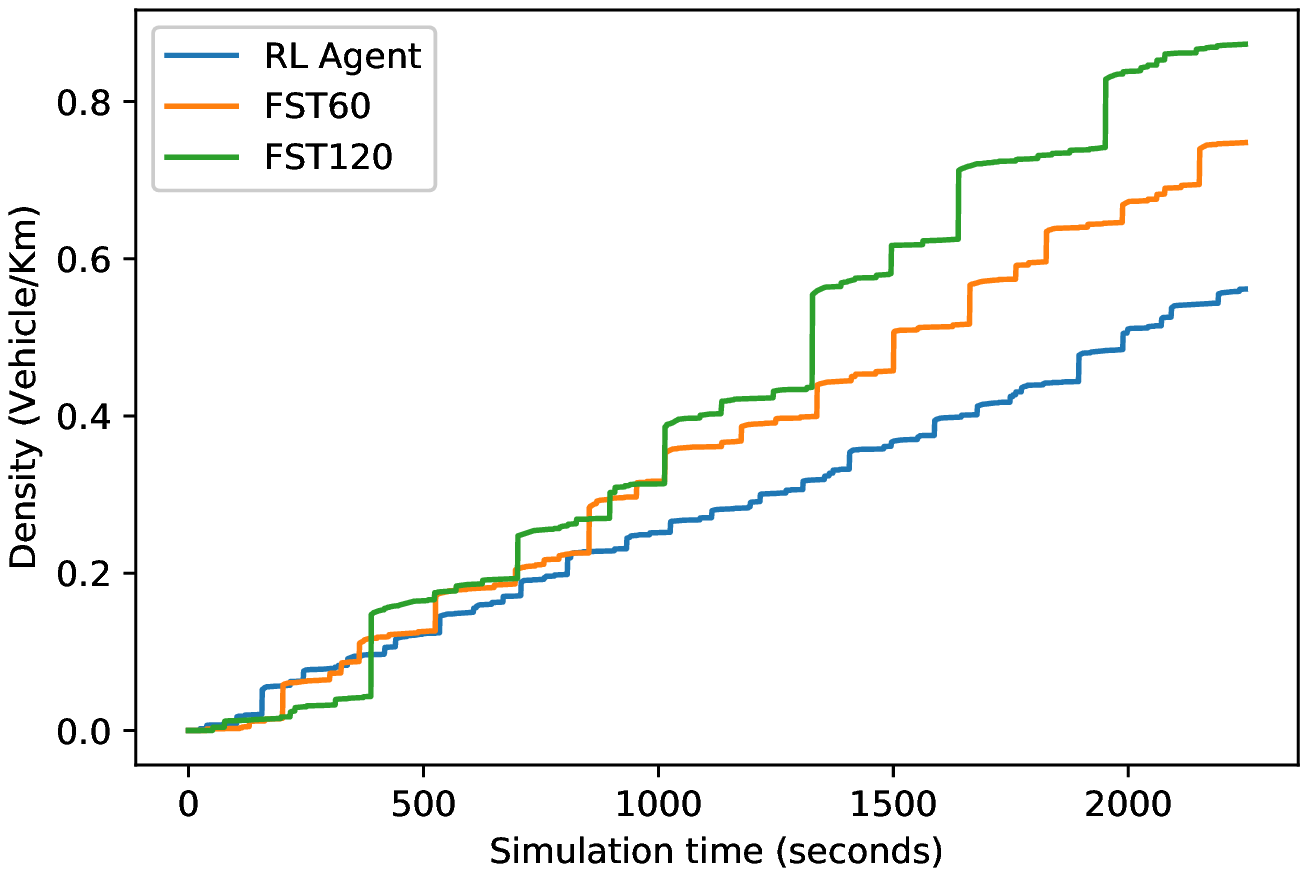}}
 
     \caption{ Plots of single agent showing comparisons between FST 60, 120 and our RL agent. On y-axis is average delay time of the intersection in seconds per kilometer in (a) and average density of the intersection in vehicle per kilometer in (b). Both figures have simulation time in seconds on x-axis.}
     \label{fig:single agent graphs}
\end{figure}
\begin{figure}
     \centering
     
     \subfloat[][A1:Async A3C\\ B1:Async A3C+Coordination]{\includegraphics[width=.5\textwidth]{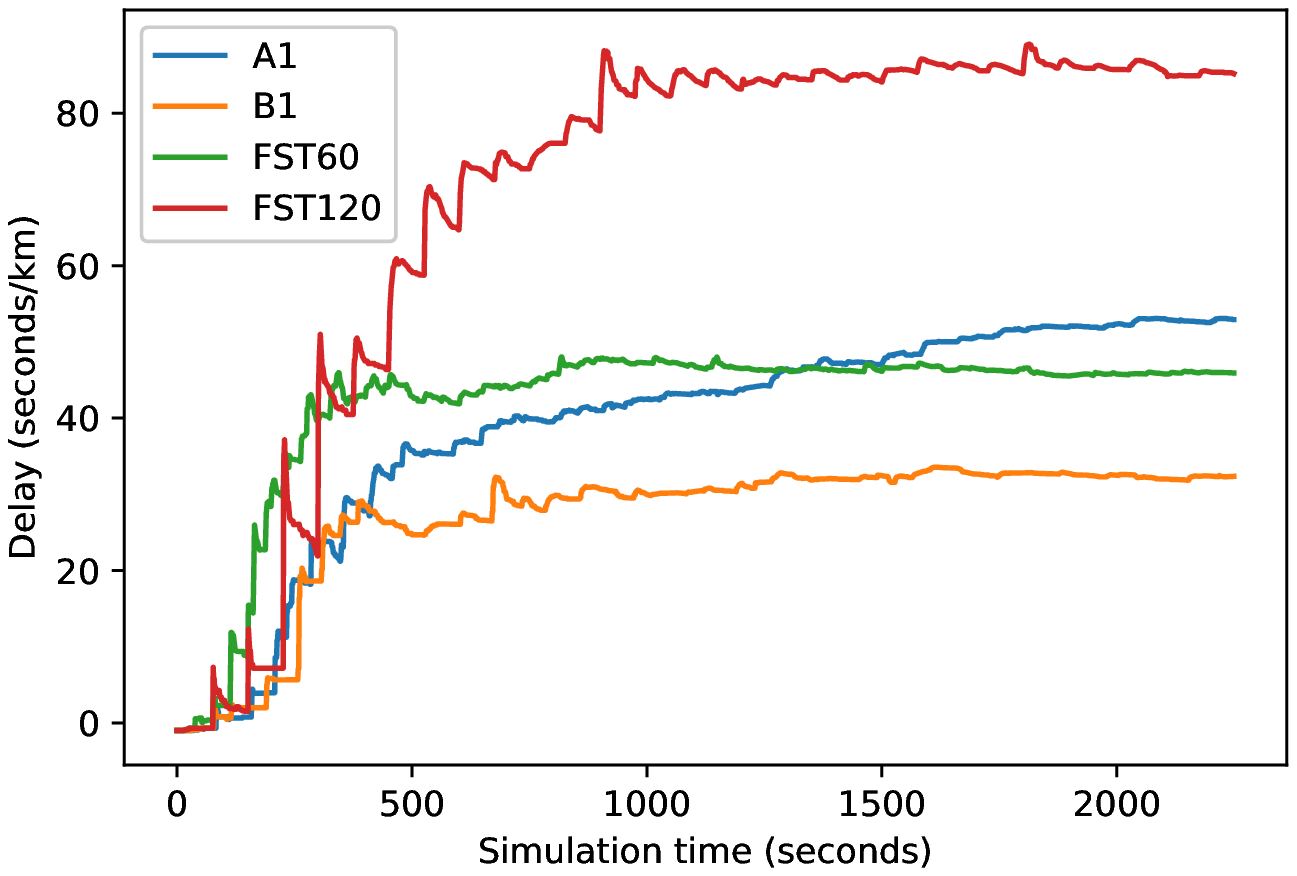}}
     \subfloat[][A2:InRL\\ B2:InRL+Coordination]{\includegraphics[width=.5\textwidth]{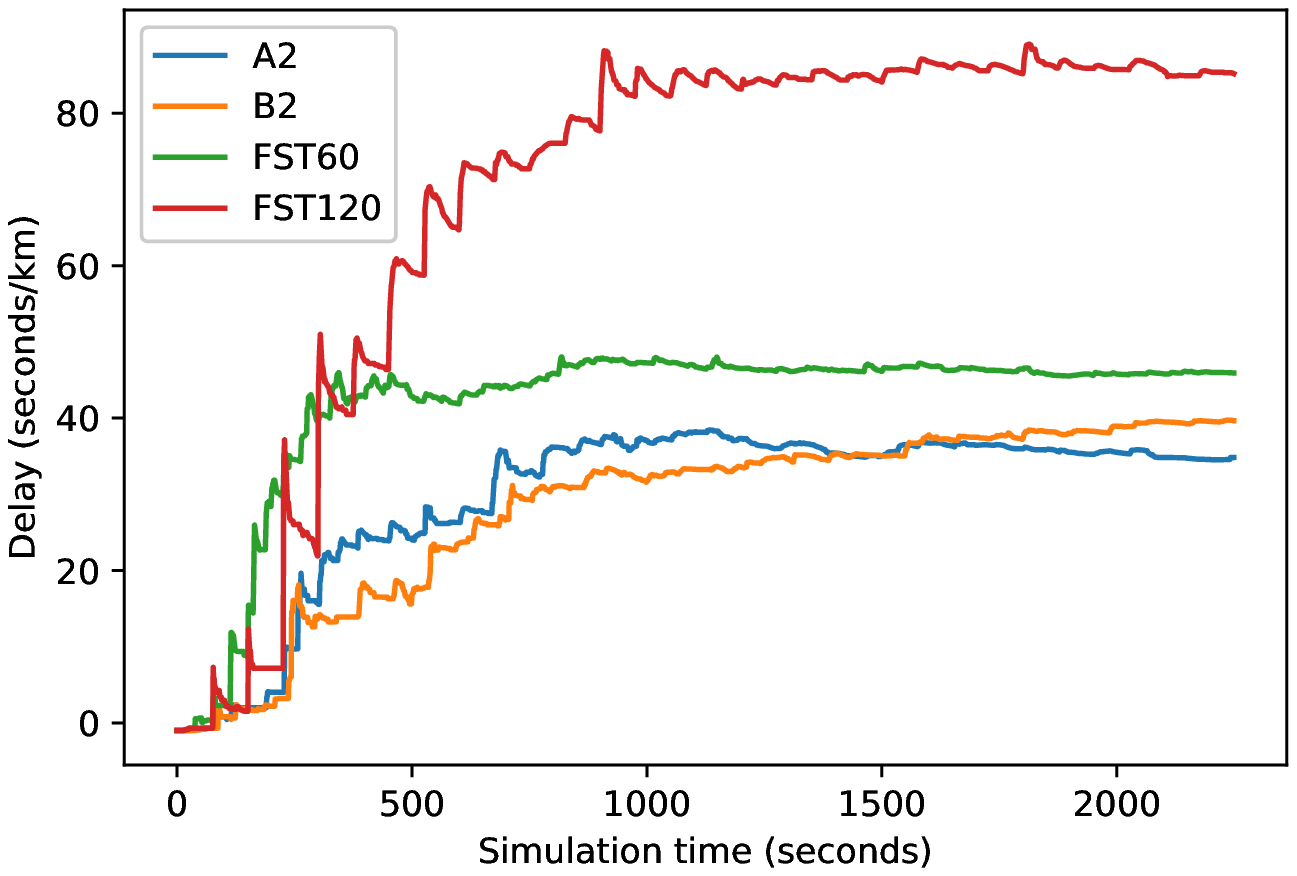}}
     \caption{Delay graphs with coordination i.e global reward function introduced in A3C and Independent RL (InRL) on four intersections. Y-axis is average delay time of the four intersections in seconds per kilometer and x-axis represents the simulation time in seconds. }
     \label{fig:Multi agent graphs}
\end{figure}

\section{Results and Conclusions} \label{result}
We describe the results of single and multi-agent RL experiments in this section. 

Fig.\ref{fig:single agent graphs} compares the average delay time and average density achieved by a single agent with FST 60, FST 90 and FST 120 at an intersection. The RL agent reduces the average density and delay by 33\% compared to FST 60 and 66.67\% compared to FST 120 respectively. This suggests that the drop in delay is because the RL agent, unlike fixed signal timing, considers real-time traffic density and dynamically adapts to the changing conditions. 

The results of experiments with the multi-agent setup, shown in Fig.\ref{fig:Multi agent graphs} lead to similar conclusions where RL performs better than FST in reducing the average delay time. It is seen in Fig.\ref{fig:Multi agent graphs}(a) that although the Async A3C algorithm is successful in lowering the delay at intersections initially, it starts performing similar to FST 60 after ~\textasciitilde1500 simulation seconds. With the introduction of the global reward function, the delay across intersections dropped by 38\%. The results confirm that coordination among agents along with Async A3C results in lower congestion across neighbouring intersections. 

Fig.\ref{fig:Multi agent graphs}(b) shows a drop in average delay in a multi-agent setup when agents are trained independent of state information of other agents. InRL achieved better results than Async A3C  because each agent focused on reducing the delay at its intersection and learnt strategies to coordinate with each other to optimize globally. It is observed that while using this technique the green time at intersections synced in a manner to achieve a coordinated green. On introducing a global reward function in the setup, a drop in delay is seen initially and at ~\textasciitilde1500 simulation seconds the average delay is similar to FST 60. 

In this paper, we evaluated multiple methods to optimize green signal time at single intersections and across multiple intersections using deep reinforcement learning. Using A3C \cite{mnih2016asynchronous} algorithm and multi-agent state space we introduced information sharing and coordination between intersections, further reducing the delay time. The experiments demonstrated that training different agents independently in a multi-agent setting led to self competitive behaviour among them, thus working better than fixed signal timing. Finally, introducing a global reward function in both Async A3C and InRL methods induced team work and cooperation among agents and their respective intersections for a smoother traffic flow. Our RL agents consider real-time traffic densities and show better performance than fixed signal timing in chaotic traffic conditions. Directions for future work include introducing traffic anomalies and lane prioritization.
\medskip

\section{Acknowledgements}

The authors acknowledge the support provided by Professor Shalabh Bhatnagar and Jayanth Prakash Kulkarni, Department of Computer Science and Automation (CSA), Indian Institute of Science Bengaluru. Their expertise and recommendations assisted the research conducted on adaptive signal control with Reinforcement Learning techniques, especially the use of Asynchronous Advantage Actor Critic (A3C) algorithm and Aimsun simulation environment.

\bibliographystyle{unsrt}
\bibliography{bibliography}

\end{document}